%% file: main.tex
\documentclass{article}

\usepackage{microtype}
\usepackage{graphicx}
\usepackage{subfigure}
\usepackage{booktabs} 

\usepackage{times}
\usepackage{epsfig}
\usepackage{graphicx}
\usepackage{amsmath}
\usepackage{amssymb}
\usepackage{dblfloatfix}

\usepackage{comment}
\usepackage{color}
\usepackage{placeins}
\usepackage{enumitem}
\usepackage{multirow}
\usepackage{dsfont}

\DeclareMathOperator*{\argmin}{arg\,min}

\newcommand{\fig}[1]{Figure~\ref{fig:#1}}
\newcommand{\sect}[1]{Section~\ref{sect:#1}}
\newcommand{\tab}[1]{Table~\ref{tab:#1}}

\usepackage{hyperref}

\usepackage[accepted]{icml2021}

\icmltitlerunning{Object Segmentation Without Labels with Large-Scale Generative Models}

\begin{document}

\twocolumn[
\icmltitle{Object Segmentation Without Labels with Large-Scale Generative Models}



\icmlsetsymbol{equal}{*}

\begin{icmlauthorlist}
\icmlauthor{Andrey Voynov}{yndx}
\icmlauthor{Stanislav Morozov}{yndx}
\icmlauthor{Artem Babenko}{yndx}
\end{icmlauthorlist}

\icmlaffiliation{yndx}{Yandex, Moscow, Russia}
\icmlcorrespondingauthor{Andrey Voynov}{an.voynov@yandex.ru}

\icmlkeywords{Machine Learning, ICML}

\vskip 0.3in
]



\printAffiliationsAndNotice{}  

\begin{abstract}
The recent rise of unsupervised and self-supervised learning has dramatically reduced the dependency on labeled data, providing effective image representations for transfer to downstream vision tasks. Furthermore, recent works employed these representations in a fully unsupervised setup for image classification, reducing the need for human labels on the fine-tuning stage as well. 
This work demonstrates that large-scale unsupervised models can also perform a more challenging object segmentation task, requiring neither pixel-level nor image-level labeling. Namely, we show that recent unsupervised GANs allow to differentiate between foreground/background pixels, providing high-quality saliency masks. By extensive comparison on standard benchmarks, we outperform existing unsupervised alternatives for object segmentation, achieving new state-of-the-art. Our model and implementation are available online\footnote[2]{\url{https://github.com/anvoynov/BigGANsAreWatching}}.
\end{abstract}

\input{intro}
\input{related}
\input{method}
\input{experiments}
\input{conclusion}

\bibliography{main}
\bibliographystyle{icml2021}

\end{document}

%% file: intro.tex
\section{Introduction}
\label{sect:intro}

Reducing the reliance on labeled data is a long-standing goal of machine learning research. The recent studies on unsupervised and self-supervised learning for both discriminative \cite{chen2020simple,he2020momentum,caron2020unsupervised,grill2020bootstrap} and generative \cite{donahue2019large,chen2020generative} models have demonstrated that one does not need labeled data on the pretraining stage to produce image representations for typical computer vision problems. While these representations require some labeled data to finetune to a particular downstream task, recent works \cite{van2020scan, zheltonozhskii2020self} exploit these representations to solve the image classification~problem~without~labels~at~all. 

This paper employs state-of-the-art unsupervised generative models to perform label-free object segmentation, where groundtruth pixel-level labels are expensive to collect because of labor-intensive human efforts. This problem currently receives much research attention and is typically addressed by methods based on GANs \cite{chen2019unsupervised, bielski2019emergence, benny2019onegan}. However, training high-quality GANs can be both time-consuming and unstable. Moreover, the protocols in \cite{chen2019unsupervised, bielski2019emergence, benny2019onegan} typically include a large number of hyperparameters that are tricky to tune in the completely unsupervised setup when a labeled validation set is not available. In contrast, we propose an alternative, much simpler approach that does not require adversarial training or heavy hyperparameter tuning for each particular segmentation task. 

Our work is partially inspired by the findings from \citet{voynov_icml_2020}, which has shown that the latent space of BigGAN \citep{brock2018large} possess the direction ``responsible'' for the background removal, and this direction can be used to produce training data for saliency detection. However, the approach \citep{voynov_icml_2020} is not unsupervised since (i) BigGAN is trained with known Imagenet labels, therefore, it is not an unsupervised model; (ii) it requires manual inspecting of several latent transformations.

This paper eliminates external supervision mentioned above and demonstrates that large off-the-shelf GANs can segment images, being completely unsupervised. As a main technical novelty, we introduce an automatic procedure identifying the ``segmenting'' latent directions in the pretrained GANs. This procedure reveals such directions in the state-of-the-art publicly available BigBiGAN \citep{donahue2019large}, which is trained on the Imagenet \citep{imagenet_cvpr09} without labels. These directions allow to distinguish object/background pixels in the generated images, providing decent segmentation masks. These masks are then used to supervise a discriminative U-Net model \citep{ronneberger2015u}, which is stable and easy to train. As another advantage, our approach also provides a straightforward way to tune hyperparameters. Since an amount of synthetic data is unlimited, its hold-out subset can be used as validation.

Our work confirms the promise of using GANs to produce synthetic training data, which is a long-standing goal of research on generative modeling. In extensive experiments, we show that the approach often outperforms the existing unsupervised alternatives for object segmentation and saliency detection. Our results provide additional evidence to the common trend that more data and larger models often can reduce the requirements of human labels.

Overall, the contributions of our paper are the following:
\begin{enumerate}

\item We propose to perform unsupervised object segmentation using off-the-shelf Imagenet-pretrained GANs.

\item We introduce an automatic method to identify ``segmenting'' directions in the GAN latent space.

\item We show that our method outperforms the state-of-the-art in most operating points. Given its simplicity, the method can serve as a baseline in the future.

\end{enumerate}

%% file: related.tex
\section{Related work}
\label{sect:related}

In this paper, we address the binary object segmentation problem, i.e., for each pixel, we aim to predict if it belongs to the object or the background. This problem is typically referred to as saliency detection \cite{wang2019salient} and foreground object segmentation \cite{chen2019unsupervised, bielski2019emergence, benny2019onegan}. While most prior works propose fully-supervised or weakly-supervised methods, we focus on the most challenging unsupervised setup, where only a few approaches have been developed.

\textbf{Existing unsupervised approaches.} Before the rise of deep learning models, a large number of ``shallow'' unsupervised techniques were developed \cite{zhu2014saliency,jiang2013salient,peng2016salient,cong2017co,cheng2014global,wei2012w}. These earlier techniques were mostly based on hand-crafted features and heuristics, e.g., color contrast \cite{cheng2014global}, or certain background priors \cite{wei2012w}. Often these approaches also utilize traditional computer vision routines, such as super-pixels \cite{yang2013saliency,wang2016correspondence}, object proposals \cite{guo2017video}, CRF \cite{Koltun_dense_crf}. These heuristics, however, are not completely learned from data, and the corresponding methods are inferior to the more recent ``deep'' approaches. 

Regarding unsupervised deep models, several works have recently been proposed by the saliency detection community \cite{wss,dus,sbf,nguyen2019deepusps}. Their main idea is to combine or fuse the predictions of several heuristic saliency methods, typically using them as a source of noisy groundtruth for deep CNN models. However, these methods are not entirely unsupervised since they rely on the supervised-pretrained classification or segmentation networks or utilize a limited number of labeled data \cite{FewShotSaliencyNips2020}. In contrast, in this work, we focus on the methods that do not require labeled data.

\textbf{Generative models for object segmentation.} The recent line of unsupervised methods \cite{chen2019unsupervised, bielski2019emergence} employs generative modeling to decompose the image into the object/background. In a nutshell, these methods exploit the idea that the object's location or appearance can be perturbed without affecting image realism. This inductive bias is formalized in the training protocols, which include learning of GANs. Therefore, for each new segmentation task, one has to perform adversarial learning, which can be unstable, time-consuming, and sensitive to hyperparameters. In contrast, our approaches avoid these disadvantages, being much simpler and easier to reproduce. In essence, we propose to use the ``inner knowledge'' of pretrained large-scale GANs to produce the saliency masks.

\textbf{Latent spaces of large-scale GANs.} Recent study \cite{voynov_icml_2020} has shown that the latent space of BigGAN \cite{brock2018large} can be used to obtain saliency masks for synthetic images. However, such an ability was discovered only for BigGAN trained under the supervision from the image class labels. For unconditional GANs, it was not discovered in \cite{voynov_icml_2020}, hence, it is not clear if the supervision from the class labels is necessary for the GAN latent space to distinguish between object/background pixels. This paper shows that this supervision is not necessary, contributing novel knowledge to the general trend to unsupervised learning. 

%% file: method.tex
\section{Latent Segmenters in Unsupervised GANs}
\label{sect:method}

\citet{voynov_icml_2020} has shown that the BigGAN's latent space contains a direction $h_{bg}$ responsible for a background removal: once a latent code $z$ of a generated image $G(z)$ is shifted by $h_{bg}$, the background pixels of the shifted image $G(z + h_{bg})$ become white, while the foreground ones remain almost unchanged. As we will show, the latent spaces of other large-scale GANs also have directions that have different effects on background/foreground pixels. The following section provides a principled framework to identify such ``segmenting'' directions automatically.

\subsection{Modeling a segmenting direction}

Formally, we consider a latent shift $h$ to be a \textit{segmenting direction} if there are two affine operators $A_1, A_2: \mathbb{R}^3 \to \mathbb{R}^3$ such that for each latent $z$ and pixel $G(z)_{x, y}$ we have
\begin{equation}
G(z + h)_{x, y} = A_{i(x, y)}(G(z)_{x, y}),\ i(x, y) \in \{1, 2\}
\label{A_decomp}
\end{equation}
that is, the latent shift acts on each pixel as one of the two fixed maps. Intuitively, this definition formalizes our desire that the latent segmenter should affect the background/foreground pixels differently.

Now we explain how to find these affine operators $A_1, \ A_2$ for a given latent direction $h$ from the generator $G$. In addition, we also show how to identify the directions that are the most appropriate for an object segmentation task. Given two affine operators $A_1, A_2$, for a pair of pixel intensities $c, c'$ let us define a map
\begin{equation}
    S_{A_1, A_2}(c, c') = \argmin\limits_{A \in \{A_1, A_2\}}(\|A(c) - c'\|_2)\ \cdot \ c
    \label{op_to_label}
\end{equation}
that is $S_{A_1, A_2}$ chooses an operator that maps $c$ closer to $c'$ and applies it to $c$. We extend this action to the generated images space by setting $(S_{A_1, A_2} \cdot G(z))_{x, y} = S_{A_1, A_2}(G(z)_{x, y}, G(z + h)_{x, y})$. Then we define the restoration loss:
\begin{equation}
    \mathcal{L}_h(A_1, A_2) = \mathop{\mathbb{E}}\limits_{z}\sum\limits_{x, y}\|\left(S_{A_1, A_2} \cdot G(z)\right)_{x, y} - G(z + h)_{x, y}\|_2
    \label{restoration_loss}
\end{equation}
here we sum over all pixels of an image $G(z)$. This quantity indicates how good one can approximate the map $\sigma_h: G(z) \to G(z + h)$ by choosing the optimal $A_1, A_2$ for each pixel. If this map can be represented in a form of \eqref{A_decomp}, the quantity $\mathcal{L}_h$ possesses a global minimum equal to 0.

Thus, for a given direction $h$ one can find the optimal $A_1, A_2$ by solving 
\begin{equation}
A_1, A_2 = \mathrm{argmin}_{A_1, A_2} \mathcal{L}_h(A_1, A_2)
\label{L_loss}
\end{equation}
These operators also define a binary segmentation of a generated image by assigning a label computed as 
\begin{equation}
\argmin_{i \in \{1, 2\}}\|A_i \cdot G(z)_{x, y} - G(z + h)_{x, y}\|_2
\label{class_def}
\end{equation}
for each pixel $(x, y)$.

\subsection{Exploring segmenting directions}
\label{sect:segm_direction_algorithm}
Now we explain how to identify the segmenting direction in the latent space of a given pretrained GAN. First, we find a set of interpretable directions using the technique from \cite{voynov_icml_2020} with the default hyperparameters. This results in a set of latent directions $h_1, \dots, h_{N}$. For each $h_k$ we then optimize (\ref{L_loss}) with the stochastic gradient descent. Since the number of learnable parameters is only $24$ and the loss is averaged over all image pixels, we use a small mini-batch of four images and $200$ steps of Adam optimizer with a learning rate $0.005$. The optimization converges rapidly, and we did not observe any benefits from larger batches or larger numbers of steps. Overall, this optimization takes a few minutes on the Nvidia-1080ti GPU card. Thus, for each $h_k$ we obtain a pair of affine operators $(A_1^{(h_k)}, A_2^{(h_k)})$. The optimal loss value $\mathcal{L}_{h_k}$ indicates how good a particular transform $\sigma_{h_k}$ can be approximated by two pixelwise affine operators. In practice, this ranking is not sufficient to identify directions suitable for segmentation as the transforms $\sigma_k$ may induce almost identical or a global lighting transformation with $A^{(h_k)}_1$ close to $A^{(h_k)}_2$. If so, the masking based on these operators becomes noisy and inadequate for downstream tasks. To overcome this issue, for each $h_k$ we compute the mean distance $D_k = \|A^{(h_k)}_1 \cdot G(z)_{x, y} - A^{(h_k)}_2 \cdot G(z)_{x, y}\|_2$ over the pixels of generated images. Intuitively, the direction that induces the most distant $A_1, A_2$ should produce adequate segmentation masks for synthetic images.

We apply the above approach to the Imagenet-pretrained unsupervised BigBiGAN, which parameters are available online. After using the method from \cite{voynov_icml_2020}, we extract $120$ latent directions to serve as candidates to be the latent segmenters. We also scale them by a multiplier $5$ as the unit-length latent shifts commonly induce minor image transformation leading to noisy restoration loss $\mathcal{L}_h$ optimization process. After solving the optimization problem (\ref{L_loss}) and computing $D_k$ values for each $h$, we choose the direction with the highest $D_k$ among the best-$70\%$ in terms of the restoration loss. We use this direction as a latent segmenter to produce the saliency masks. 

\begin{figure*}
    \centering
    $\vcenter{\hbox{\includegraphics[width=0.44\textwidth]{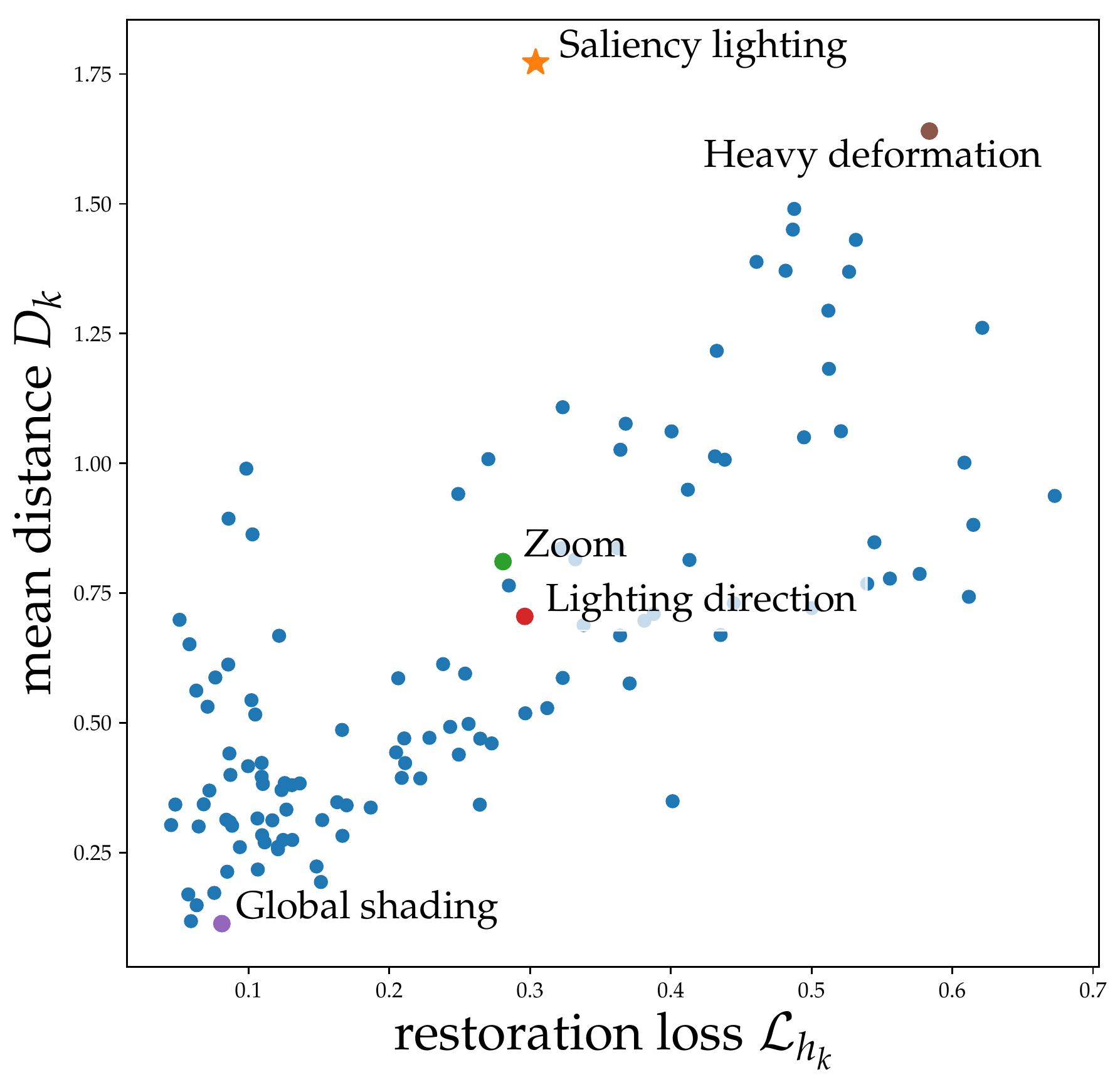}}}$
    \hspace{5mm}
    $\vcenter{\hbox{\includegraphics[width=0.43\textwidth]{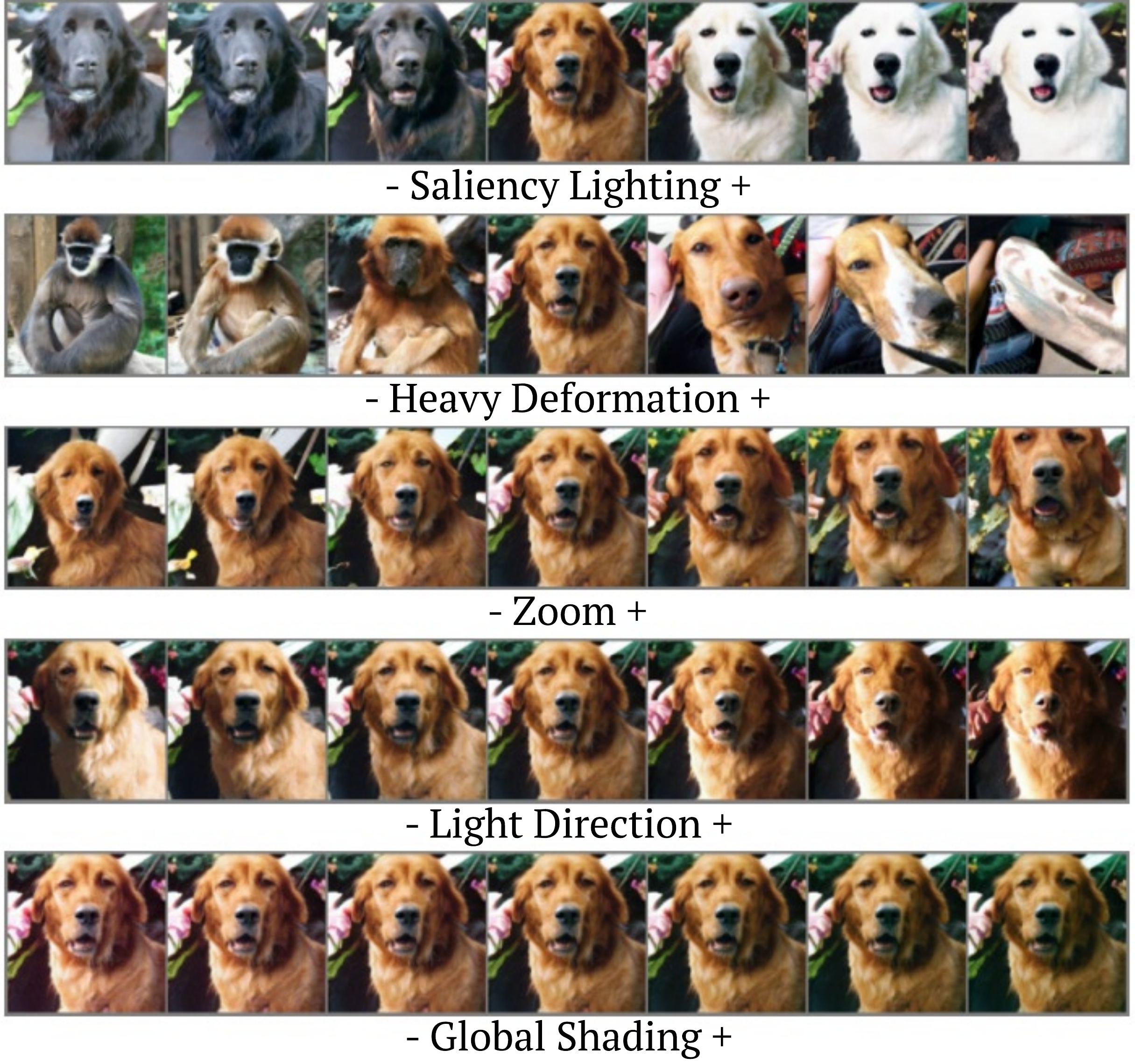}}}$
    \vspace{-4mm}
    \caption{\textit{Left}: BigBiGAN latent directions restoration loss values and the operators $A_1, A_2$ dissimilarity. \textit{Right}: examples of generated image transformations induced by moving a latent $z$ along some of directions. The central image in each row corresponds to the original sample $G(z)$.}
    \label{fig:l_d_directions_scatter}
\end{figure*}

\begin{figure}[h]
    \centering
    \includegraphics[width=\columnwidth]{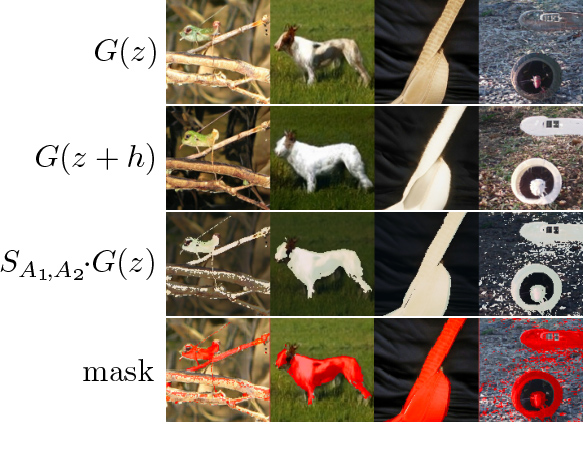}
    \vspace{-10mm}
    \caption{\textit{From top to bottom}: generated image; shifted image; shifted image approximation by pixelwise operators $A_1, A_2$; mask generated by these operators assignment.}
    \label{fig:segmentation_synth}
    \vspace{-5mm}
\end{figure}

On the \fig{l_d_directions_scatter} we plot the optimal values of the restoration loss $\mathcal{L}_{h_k}$ and the operators mean distance $D_k$ for all the candidate directions. Notably, the background/foreground direction, which we utilize for the saliency generation, has the highest mean distance $D_k$ while possessing low restoration loss $\mathcal{L}_{h_k}$. On the \fig{segmentation_synth} we illustrate the images $S_{A_1, A_2} \cdot G(z)$ that approximate the shifted $G(z + h)$ by the pixelwise operators $A_1, A_2$ that minimize $\mathcal{L}_{h_{bg}}$. In our experiments, the operators $A_1, A_2$ corresponding to the background saliency direction have a form:

\vspace{-3mm}
$$
A_1(c) = \
\begin{pmatrix}
0.13 & -0.12 & 0.06\\
0.01 & 0.00 & 0.04\\
0.02 & -0.20 & 0.22\\
\end{pmatrix} \cdot c + \
\begin{pmatrix}
0.78\\
0.76\\
0.69\\
\end{pmatrix};
$$
$$
A_2(c) = \
\begin{pmatrix}
0.31 & -0.05 & 0.05\\
0.04 & 0.19 & 0.06\\
0.01 & -0.06 & 0.31\\
\end{pmatrix} \cdot c - \
\begin{pmatrix}
0.1\\
0.15\\
0.19\\
\end{pmatrix} 
$$
\vspace{-3mm}

The first operator performs aggressive lightening, while the second one downscales the channels and applies a minor negative shift resulting in a darkening. In practice, the intensity of the pixels handled by the first operator increases while the intensity of the pixels handled by the second one decreases. This direction, along with the corresponding operators, is used to produce the saliency masks as shown in \fig{segmentation_synth}. For efficiency, we set the generated image mask at pixel $(x, y)$ to be equal to $\left[\|G(z + h_{bg})_{x, y}\| > \|G(z)_{x, y}\|\right]$ as in practice it appears to be a decent approximation of (\ref{class_def}).


\subsection{Adaptation to the particular segmentation task.}
\label{sect:adapting}

Since BigBiGAN was trained on the Imagenet, sampling the latent codes from the standard Gaussian distribution $z \sim \mathcal{N}(0,\mathbb{I})$ will result in the distribution of the synthetic data that resembles the Imagenet. However, this distribution can be suboptimal for the particular segmentation task. To mitigate this issue, we introduce a simple additional step in the process of synthetic data generation. To make the distribution of generated images closer to the particular dataset $\mathcal{I}{=}\{I_1,\dots, I_N\}$, we sample $z$ from the latent space regions that are close to the latent codes of $\mathcal{I}$. To this end, we use the BigBiGAN encoder to compute the latent representations $\{E(I_1),\dots, E(I_N)\} \subset \mathbb{R}^{120}$ and sample the codes from the neighborhood of these representations. Formally, the samples have a form:

\vspace{-3mm}
\begin{equation}
\{E(I_i){+}\alpha \xi \;|\; i\sim\mathcal{U}\{1,N\}, \xi \sim \mathcal{N}(0,I)\}
\end{equation}

Here $\alpha$ denotes the neighborhood size, and it should be larger for small $\mathcal{I}$ to prevent overfitting. In particular, we use $\alpha{=}0$ for Imagenet and $\alpha{=}0.2$ for all other cases. In the experimental section, we demonstrate that this simple and efficient modification of the data generation process results in a dramatic performance boost.


\subsection{Improving saliency masks.}
\label{sect:filtering}

Here we describe a few simple heuristics that increase the masks' quality for the particular segmentation task. The ablation of each component is presented in \sect{ablation}.

\textbf{Mask size filtering.} Since some of the BigBiGAN-produced images are low-quality and do not contain well-defined objects, the corresponding masks can result in very noisy supervision. To alleviate this, we apply simple filtering that excludes the images where the ratio of foreground pixels exceeds $0.5$.

\textbf{Histogram filtering.} Since $G(z{+}h_{bg})$ should have mostly dark and light pixels, we filter out the images that are not contrastive enough. Formally, we compute the intensity histogram with $12$ bins for the grayscaled $G(z{+}h_{bg})$. Then we smooth it by taking the moving average with a window of $3$ and filter out the samples that have local maxima outside the first/last buckets of the histogram.

\begin{figure*}[ht!]
\noindent
\centering
\vspace{-3mm}
\includegraphics[width=\textwidth]{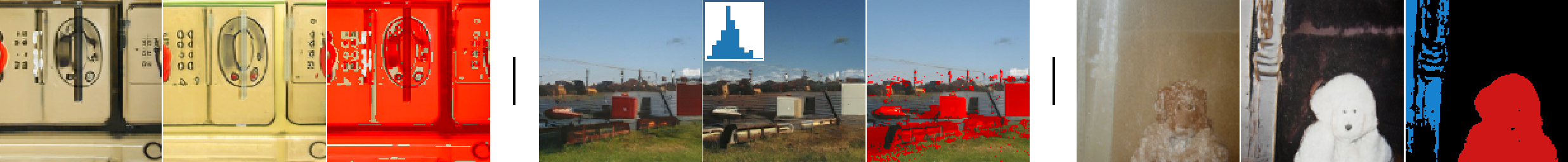}
\vspace{-5mm}
\caption{Examples of mask improvement. \textit{Left:} the sample rejected by the mask size filter. \textit{Middle:} the sample rejected by the histogram filtering. \textit{Right block}: mask pixels removed by the connected components filter are shown in blue and the remaining mask pixels are shown in red.}
\label{fig:filtering}
\end{figure*}

\begin{figure*}[b!]
\noindent
\centering
\includegraphics[width=0.9\textwidth]{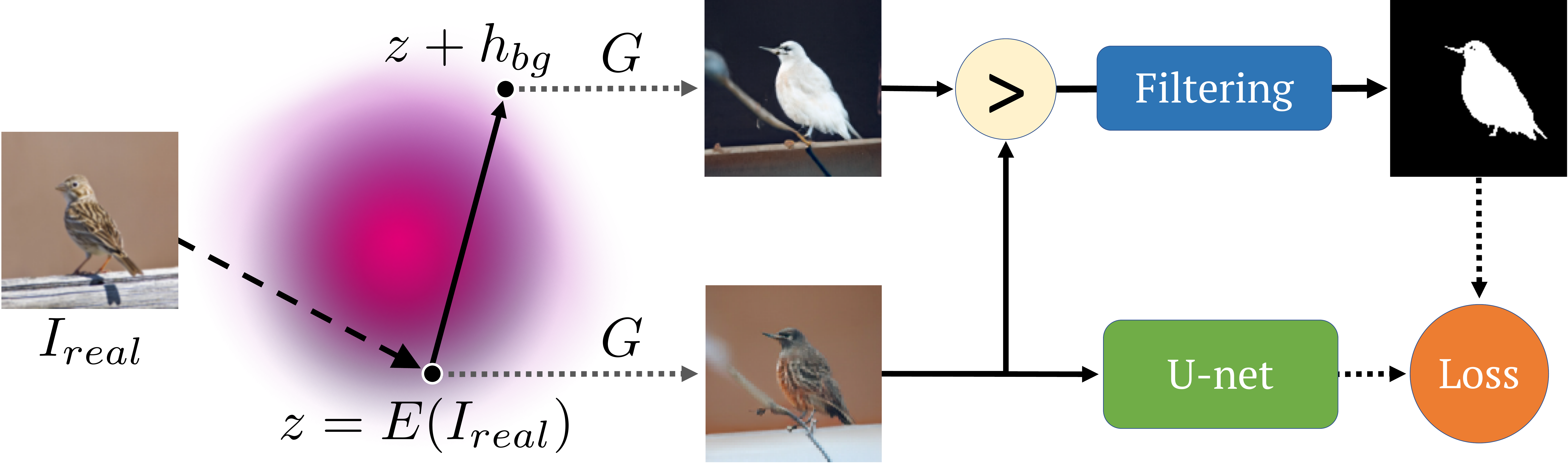}
\vspace{-3mm}
\caption{Schematic representation of our approach. First, we generate an image $G(z)$ and the shifted image $G(z + h_{bg})$. Then these images induce a synthetic mask. This mask is passed through a filtering (\sect{filtering}) and a U-net segmentation model learns to predict it, given the original generated image.}
\label{fig:scheme}
\end{figure*}

\textbf{Connected components filtering.} For each generated mask $M$ we group the foreground pixels into connected (by edges) groups forming clusters $M_1, \dots, M_k$. Assuming that $M_1$ is the cluster with the maximal area, we exclude all the clusters $M_i$ with $|M_i| < 0.2 \cdot |M_1|$. This technique allows to remove visual artifacts from the synthetic data. We present samples of images rejected by each filtering step in \fig{filtering}.

\subsection{Training the model on synthetic data}
\label{sect:synth_train_details}

\begin{figure}[h!]
    \centering
    \includegraphics[width=0.49\columnwidth]{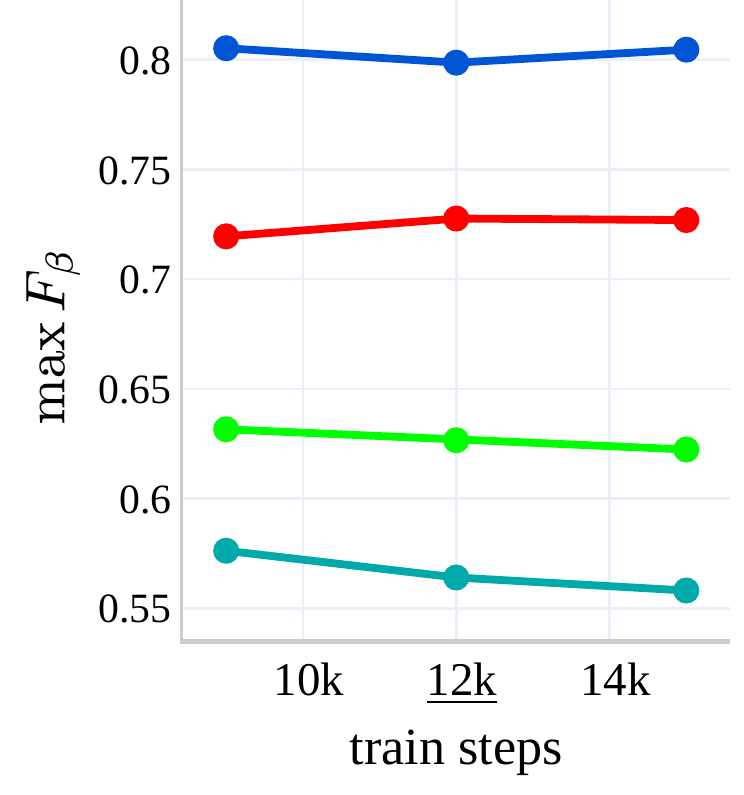}    \includegraphics[width=0.49\columnwidth]{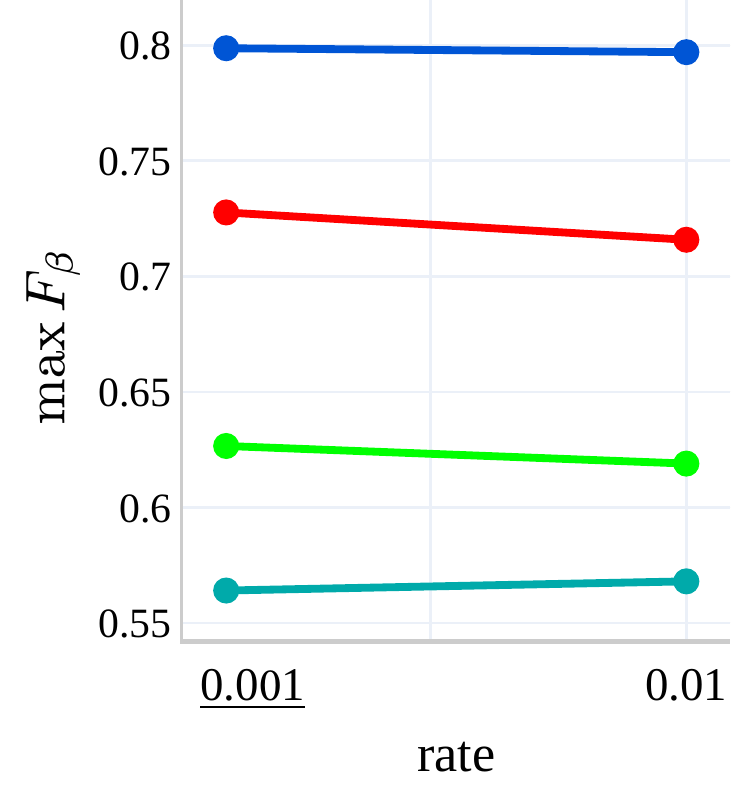}
    \includegraphics[width=0.49\columnwidth]{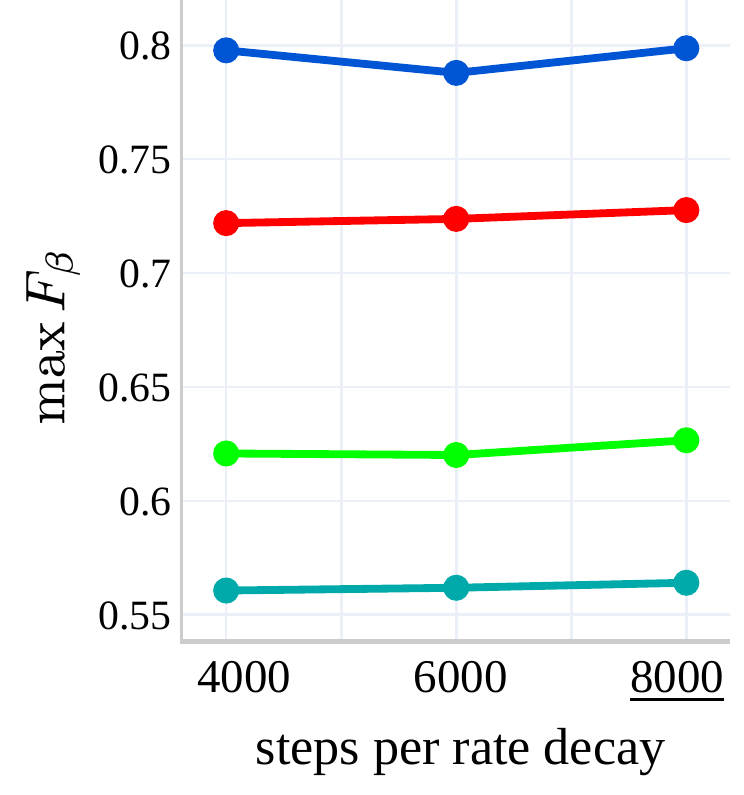}
    \includegraphics[width=0.49\columnwidth]{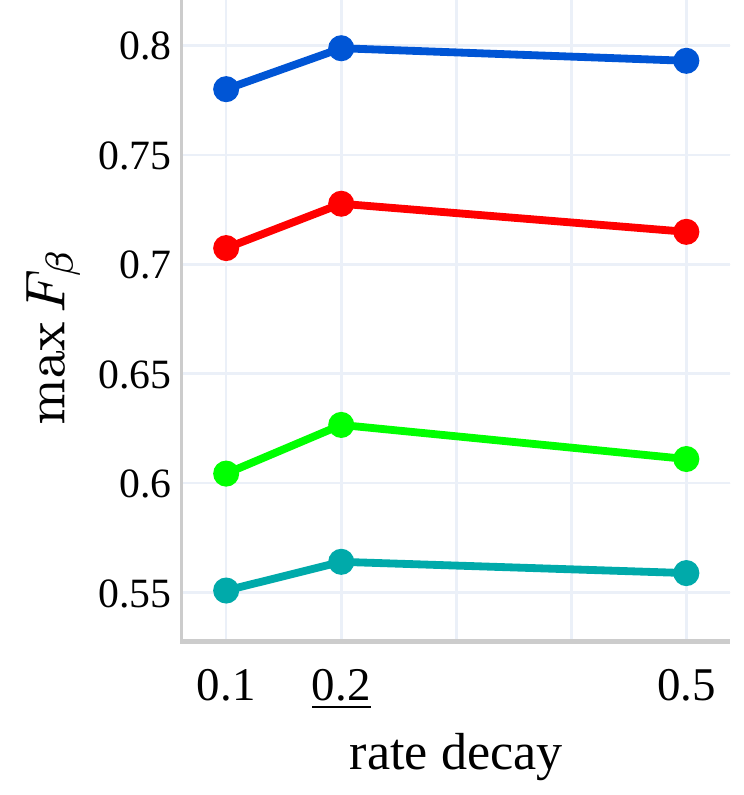}

    \vspace{4mm}

    \includegraphics[width=0.49\columnwidth]{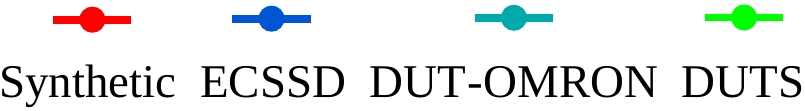}
    \caption{Model performance for different hyperparameters. We take the hyperparameters that performs best on the synthetic data.}
    \label{fig:synth_hypperparam_tune}
\end{figure}

Given a large amount of synthetic data, one can train one of the existing image-to-image CNN architectures in the fully supervised regime. The whole pipeline is schematically presented in \fig{scheme}. In all our experiments, we employ a standard U-net architecture \cite{ronneberger2015u}. We train U-net on the synthetic dataset with the Adam optimizer and the binary cross-entropy objective applied on the pixel level. We perform $12 \cdot 10^3$ steps with batch 95. The initial learning rate equals $0.001$ and is decreased by $0.2$ on step $8 \cdot 10^3$. During inference, we rescale an input image to have a size 128 along its shorter side. Compared to existing unsupervised alternatives, the training of our model is straightforward and does not include a large number of hyperparameters. The only hyperparameters in our protocol are batch size, learning rate schedule, and a number of optimizer steps, and we tune them on the hold-out validation set of synthetic data. We set the batch size to guarantee the maximal GPU memory utilization. \fig{synth_hypperparam_tune} reports the segmentation quality on the hold-out synthetic data for the different hyperparameter values. Notably, they have a minor affect on the model quality. \fig{synth_hypperparam_tune} demonstrates that the optimal hyperparameters chosen for synthetic data are typically optimal for real datasets as well. Training with online synthetic data generation takes approximately seven hours on two Nvidia 1080Ti cards.

\hspace{3mm}

\input{segm_table}

%% file: segm_table.tex
\begin{table*}[b]
 \centering
 \renewcommand\arraystretch{1.2}
\setlength{\tabcolsep}{2pt}
\begin{tabular}{|c|c|c|c|c|c|c|}
\hline
\multirow{2}{*}{Method} &\multicolumn{3}{|c|}{CUB-200-2011}&\multicolumn{3}{|c|}{Flowers}\\
\cline{2-7}
 & max $F_{\beta}$ & IoU & Accuracy & max$F_{\beta}$ & IoU & Accuracy\\
\hline
PerturbGAN & --- & 0.380 & --- & --- & --- & ---\\
ReDO & --- & 0.426 & 0.845 & --- & 0.764 & 0.879\\
OneGAN & --- & 0.555 & --- & --- & --- & ---\\
BigBiGAN & 0.794 & 0.683 & 0.930 & 0.760 & 0.540 & 0.765\\
E-BigBiGAN (w/o $z$-noising) & 0.750 & 0.619 & 0.918 & 0.814 & 0.689 & 0.874\\
E-BigBiGAN (with $z$-noising) & \textbf{0.834} & \textbf{0.710} & \textbf{0.940} & \textbf{0.878} & \textbf{0.804} & \textbf{0.904}\\
std & 0.005 & 0.007 & 0.002 & 0.001 & $<$0.001 & $<$0.001\\
\hline
\end{tabular}
\caption{The comparison of unsupervised object segmentation methods. For our model, we report the performance averaged over ten runs. For the best model, we also report the standard deviation values.}

\label{tab:segmentation}
\end{table*}

%% file: experiments.tex
\section{Experiments}
\label{sect:experiments}

This section aims to confirm that the usage of GAN-produced synthetic data is a promising direction for unsupervised saliency detection and object segmentation. To this end, we extensively compare our approach to the existing unsupervised counterparts on the standard benchmarks.

\input{main_table}

\textbf{Evaluation metrics.} All the methods are compared in terms of the three measures described below.

\begin{itemize}
    \item \textbf{F-measure} is an established measure in the saliency detection literature. It is defined as $F_{\beta}{=}\frac{(1+\beta^2){Precision}\times{Recall}}{\beta^2{Precision}{+}{Recall}}$.
    Here Precision and Recall are calculated based on the binarized predicted masks and groundtruth masks as ${Precision}{=}\frac{TP}{TP{+}FP}$ and ${Recall}{=}\frac{TP}{TP{+}FN}$, where TP, TN, FP, FN denote true-positive, true-negative, false-positive, and false-negative, respectively. We compute F-measure for 255 uniformly distributed binarization thresholds and report its maximum value ${max}$ $F_{\beta}$. We use $\beta{=}0.3$ for consistency with existing works.

    \item \textbf{IoU} (Intersection over Union) is calculated on the binarized predicted masks and groundtruth as ${IoU}(s, m){=}\dfrac{\mu (s \cap m)}{\mu (s \cup m)}$, where $\mu$ denotes the area. The binarization threshold is set to $0.5$.

    \item \textbf{Accuracy} measures the proportion of pixels that have been correctly assigned to the object/background. The binarization threshold for masks is set to $0.5$. 
\end{itemize}

Since the existing literature uses different benchmark datasets for saliency detection and object segmentation, we perform a separate comparison for each task below.

\subsection{Object segmentation.}

\textbf{Datasets.} We use two following datasets from the literature of segmentation with generative models.

\begin{itemize}
    \item \textbf{Caltech-UCSD Birds 200-2011} \cite{wah2011caltech} contains 11,788 photographs of birds with segmentation masks. We follow \cite{chen2019unsupervised}, and use 10,000 images for our training subset and 1,000 for the test subset from splits provided by \cite{chen2019unsupervised}. Unlike \cite{chen2019unsupervised}, we do not use any images for validation and simply omit the remaining 788 images.
    \item \textbf{Flowers} \cite{nilsback2007delving} contains 8,189 images of flowers equipped with saliency masks generated automatically via the method developed for flowers. We do not apply the mask area filter in our method with this dataset, as it rejects most of the samples.
\end{itemize}

\begin{figure*}[b]
\noindent
\centering
\includegraphics[width=0.95\textwidth]{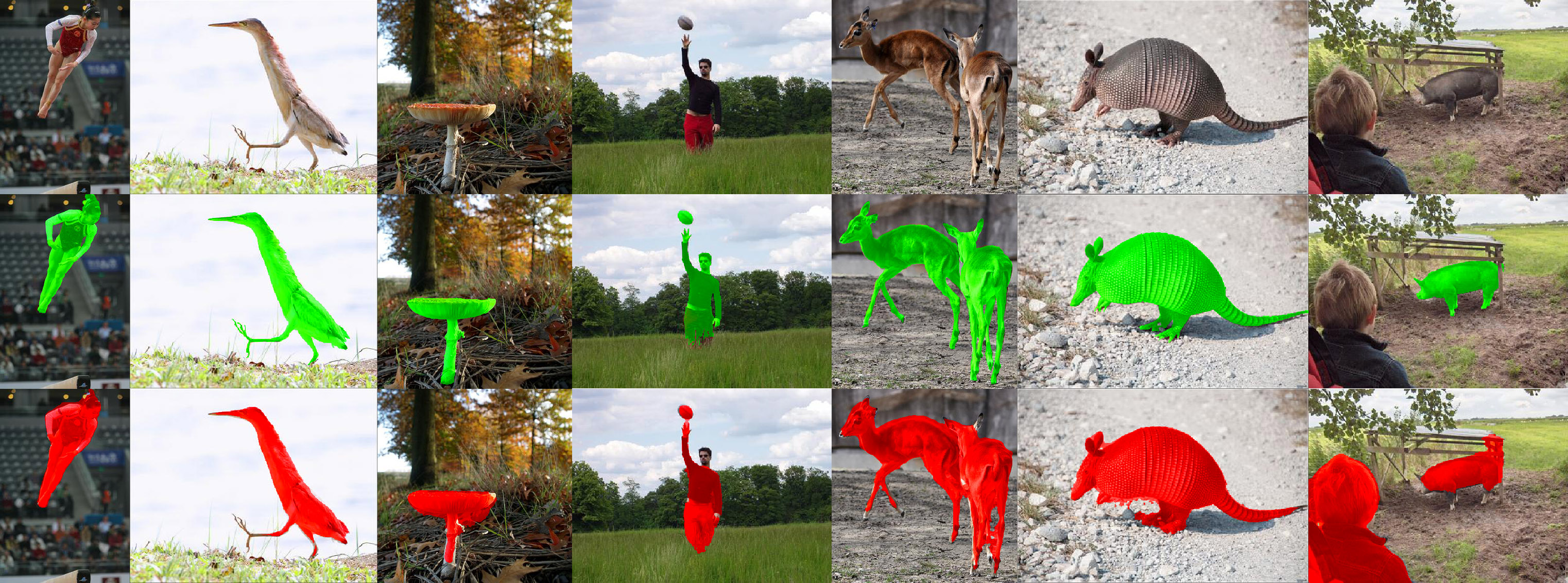}
\caption{\textit{Top:} Images from the DUTS-test dataset. \textit{Middle:} Groundtruth masks. \textit{Bottom:} Masks produced by the E-BigBiGAN method.}
\label{fig:duts_samples}
\end{figure*}

On these two datasets, we compare the following methods:

\begin{itemize}
    \item \textbf{PerturbGAN} \cite{bielski2019emergence} segments an image based on the idea that object location can be perturbed without affecting the scene realism. For comparison, we use the numbers reported in \cite{bielski2019emergence}.
    
    \item \textbf{ReDO} \cite{chen2019unsupervised} produces segmentation masks based on the idea that object appearance can be changed without affecting image quality. For comparison, we report the numbers from \cite{chen2019unsupervised}. Note, \cite{chen2019unsupervised} use hold-out labeled sets to set hyperparameters.
    
    \item \textbf{OneGAN} \cite{benny2019onegan} which simultaneously learning a conditional image generator, foreground extraction and segmentation, clustering, and object removal and background completion. Note that \cite{benny2019onegan} uses bounding boxes from image annotations to cut background patches, which is a source of weak supervision.
    
    \item \textbf{BigBiGAN} is our method where the latent codes are sampled from $z \sim \mathcal{N}(0,\mathbb{I})$. For Flowers dataset we found it beneficial to generate the saliency masks by thresholding the shifted image $G(z + h_{bg})$ with its mean value. Thus, for Flowers the masks are generated as $M = [G(z + h_{bg}) > \mathrm{mean}(G(z + h_{bg}))]$.
    
    \item \textbf{E-BigBiGAN (w/o $z$-noising)} is our method where the latent codes of synthetic data are sampled from the outputs of the encoder $E$ applied to the train images of the dataset at hand.
    
    \item \textbf{E-BigBiGAN (with $z$-noising)} same as above with latent codes sampled from the vicinity of the embeddings with the neighborhood size $\alpha$ set to $0.2$.
\end{itemize}

Following the prior works, we apply image preprocessing by extracting a central crop and resizing it to $128 \times 128$. The comparison results are provided in \tab{segmentation}, which demonstrates the significant advantage of our scheme. Note, since both datasets in this comparison are small-scale, $z$-noising considerably improves the performance, increasing the diversity of training images.


\begin{table*}[h]
 \centering
 \renewcommand\arraystretch{1.3}
\setlength{\tabcolsep}{1.6pt}
\begin{tabular}{|c|c|c|c|c|c|c|c|c|c|}
\hline
\multirow{2}{*}{Method} &\multicolumn{3}{|c|}{ECSSD}&\multicolumn{3}{|c|}{DUTS}&\multicolumn{3}{|c|}{DUT-OMRON}\\
\cline{2-10}
& ${max} F_{\beta}$ & IoU & Accuracy & ${max} F_{\beta}$ & IoU & Accuracy & ${max} F_{\beta}$ & IoU & Accuracy\\
\hline
\cite{voynov_icml_2020} & 0.778 & 0.648 & 0.904 & 0.604 & 0.478 & \textbf{0.889} & 0.56 & 0.444 & \textbf{0.878}\\
BigBiGAN (base) & 0.737 & 0.626 & 0.859 & 0.575 & 0.454 & 0.817 & 0.498 & 0.389 & 0.758\\
E-BigBiGAN & \textbf{0.797} & \textbf{0.684} & \textbf{0.906} & \textbf{0.624} & \textbf{0.511} & 0.882 & \textbf{0.563} & \textbf{0.464} & 0.860\\

\hline
\end{tabular}
\caption{Comparison of our method with the weakly-supervised BigGAN-based approach.}

\label{tab:biggan_vs_bigbigan}

\end{table*}

\begin{table*}[!ht]
 \centering
 \renewcommand\arraystretch{1.3}
\setlength{\tabcolsep}{1.6pt}
\begin{tabular}{|c|c|c|c|c|c|c|c|c|c|}
\hline
\multirow{2}{*}{Method} &\multicolumn{3}{|c|}{ECSSD}&\multicolumn{3}{|c|}{DUTS}&\multicolumn{3}{|c|}{DUT-OMRON}\\
\cline{2-10}
& $\text{max} F_{\beta}$ & IoU & Accuracy & $\text{max} F_{\beta}$ & IoU & Accuracy & $\text{max} F_{\beta}$ & IoU & Accuracy\\
\hline
Base & 0.737 & 0.626 & 0.859 & 0.575 & 0.454 & 0.817 & 0.498 & 0.389 & 0.758\\
+Imagenet embeddings & 0.773 & 0.657 & 0.874 & 0.616 & 0.483 & 0.832 & 0.533 & 0.413 & 0.772\\
+Size filter & 0.781 & 0.670 & 0.900 & 0.62 & 0.499 & 0.871 & 0.552 & 0.443 & 0.842\\
+Histogram & 0.779 & 0.670 & 0.900 & 0.621 & 0.503 & 0.875 & 0.555 & 0.450 & 0.850\\
+Connected components & \textbf{0.797} & \textbf{0.684} & \textbf{0.906} & \textbf{0.624} & \textbf{0.511} & \textbf{0.882} & \textbf{0.563} & \textbf{0.464} & \textbf{0.860}\\

\hline
\end{tabular}
\caption{Impact of different components in the E-BigBiGAN pipeline.}

\label{tab:ablation}
\end{table*}

\begin{table*}[h]
 \centering
 \renewcommand\arraystretch{1.3}
\setlength{\tabcolsep}{1.6pt}
\begin{tabular}{|c|c|c|c|c|c|c|c|c|c|}
\hline
\multirow{2}{*}{Method} &\multicolumn{3}{|c|}{ECSSD}&\multicolumn{3}{|c|}{DUTS}&\multicolumn{3}{|c|}{DUT-OMRON}\\
\cline{2-10}
& $\text{max} F_{\beta}$ & IoU & Accuracy & $\text{max} F_{\beta}$ & IoU & Accuracy & $\text{max} F_{\beta}$ & IoU & Accuracy\\
\hline
E-BigBiGAN with $\mathrm{U^2Net}$ & \textbf{0.813} & 0.674 & \textbf{0.911} & \textbf{0.654} & \textbf{0.525} & \textbf{0.906} & \textbf{0.663} & \textbf{0.559} & \textbf{0.915}\\
E-BigBiGAN & 0.797 & \textbf{0.684} & 0.906 & 0.624 & 0.511 & 0.882 & 0.563 & 0.464 & 0.860\\
\hline
\end{tabular}

\caption{Comparison of masks generation with supervised $\mathrm{U^2Net}$-guided synthetic labeling}

\label{tab:vs_u2net}
\end{table*}

\subsection{Saliency detection.}

\textbf{Datasets.} We use the following established benchmarks for saliency detection. For all the datasets, groundtruth pixel-level saliency masks are available. 

\begin{itemize}
    \item \textbf{ECSSD} \cite{shi2015hierarchical} contains 1,000 images with structurally complex natural contents. 
    \item \textbf{DUTS} \cite{wang2017learning} contains 10,553 train and 5,019 test images. The train images are selected from the ImageNet detection train/val set. The test images are selected from the ImageNet test, and the SUN dataset \cite{xiao2010sun}. We always report the performance on the DUTS-test subset.
    \item \textbf{DUT-OMRON} \cite{yang2013saliency} contains 5,168 images of high content variety.
\end{itemize}

\textbf{Baselines.} While there are a large number of papers on unsupervised deep saliency detection, all of them employ pretrained supervised models in their training protocols. For example, DeepUSPS \cite{nguyen2019deepusps} uses a segmentation model pretrained on CityScapes dataset \cite{cordts2016cityscapes}. Therefore, we mostly use the recent ``shallow'' methods HS \cite{HS}, wCtr \cite{wCtr}, and WSC \cite{WSC} as the baselines. These three methods were chosen based on their state-of-the-art performance reported in the literature and publicly available implementations. To compare with deep saliency detection models, we also add DeepUSPS \cite{nguyen2019deepusps} to the list of baselines. However, to perform a fair comparison, we train the DeepUSPS model without pretraining on the CityScapes dataset. The results for all methods are reported in \tab{saliency}. In this table, BigBiGAN denotes the version of our method where the latent codes of synthetic images are sampled from $z \sim \mathcal{N}(0,\mathbb{I})$. In turn, in E-BigBiGAN, $z$ are sampled from the latent codes of Imagenet-train images for all three datasets. Since the Imagenet dataset is large enough, we do not employ $z$-noising in this comparison.

As one can see, our method mostly outperforms the competitors by a considerable margin, which confirms the promise of using synthetic imagery in unsupervised scenarios. Note that DeepUSPS shows weak results using the same amount of supervision as our method. Several qualitative segmentation samples are provided on \fig{duts_samples}.


\subsection{Is BigGAN's supervision necessary for the segmentation performance?}
\label{sect:biggan_vs_bigbi}

In \tab{biggan_vs_bigbigan} we compare our method with the approach proposed in \cite{voynov_icml_2020}. Though this method is not fully unsupervised, it is interesting to compare synthetic from supervised and unsupervised GANs. \cite{voynov_icml_2020} utilized the ``background removal'' direction in the BigGAN's latent space to generate foreground / background masks. As BigGAN has no encoder, we also compare it with a weaker version of our method that uses the prior latent distribution without any filtering (see \tab{ablation}, first line). Notably, even without any adaptation to the particular dataset and filtering, our method performs on par with the ``supervised'' one. Enriched with the adaptation step, our approach outperforms \cite{voynov_icml_2020} while being unsupervised. These results are quite surprising since BigGAN has remarkably higher generation quality with the Fr\'{e}chet Inception Distance (FID) of $10.2$ facing $23.3$ for BigBiGAN.

\subsection{Ablation.}
\label{sect:ablation}

In \tab{ablation} we demonstrate the impact of individual components in our method. First, we start with a saliency detection model trained on the synthetic data pairs $\{G(z), M = \left[G(z{+}h_{bg}) > G(z)\right]\}$ with $z \sim \mathcal{N}(0,I)$. Then we add one by one the components listed in \sect{adapting} and \sect{filtering}. The most significant performance impact comes from using the latent codes of the real images from the Imagenet.

\subsection{Synthetic Data Quality}
\label{sect:gen_vs_sota}

This section compares the quality of saliency masks obtained with our method with the real ones. First, we evaluate the consistency of the real and generated masks. We use the SOTA publicly available saliency model\footnote{\url{https://github.com/NathanUA/U-2-Net}} to evaluate the quality of our synthetic masks used for the best E-BigBiGAN run with the Imagenet embeddings. The model results in $0.412$ IoU and $0.720$ accuracy on $10^5$ random samples, which is lower than our scheme's performance on the real datasets. We attribute such behavior to the fact that our synthetic data is often noisy, and the model trained on human-provided masks is not robust to the noise. In other words, it is more beneficial to train on difficult, noisy data and to test on refined, clean data rather than otherwise.

To understand whether the performance bottleneck of our method is the quality of generated images or the quality of masks, we perform the following experiment. Using the same SOTA supervised model $\mathrm{U^2Net}: \mathbb{R}^{3 \times 128 \times 128} \to \{0, 1\}^{128 \times 128}$ as above, we take randomly sampled latent $z$ formed by the E-BigBiGAN pipeline and form the dataset of the pairs $\{G(z),\ \mathrm{U^2Net}(G(z))\}$ were $G$ is the BigBiGAN generator. In other words, we take the same images as in our best method and form the masks with a high-quality saliency model pretrained on the real data.
Then we train a U-net segmentation model on this data following the protocol described in \sect{synth_train_details}. The comparison of the original model with the $\mathrm{U^2Net}$-guided model is presented in \tab{vs_u2net}. Notably, the $\mathrm{U^2Net}$-guided model performs better, though it does not demonstrate a break-through outperformance on two datasets out of three. This result indicates that the bottleneck of our method mainly lies in the quality of generated images rather than the mask generation approach.

\subsection{Independent Masks Estimation}
\label{sect:indep_masks}
\begin{figure}[h!]
    \centering
    \includegraphics[width=0.95\columnwidth]{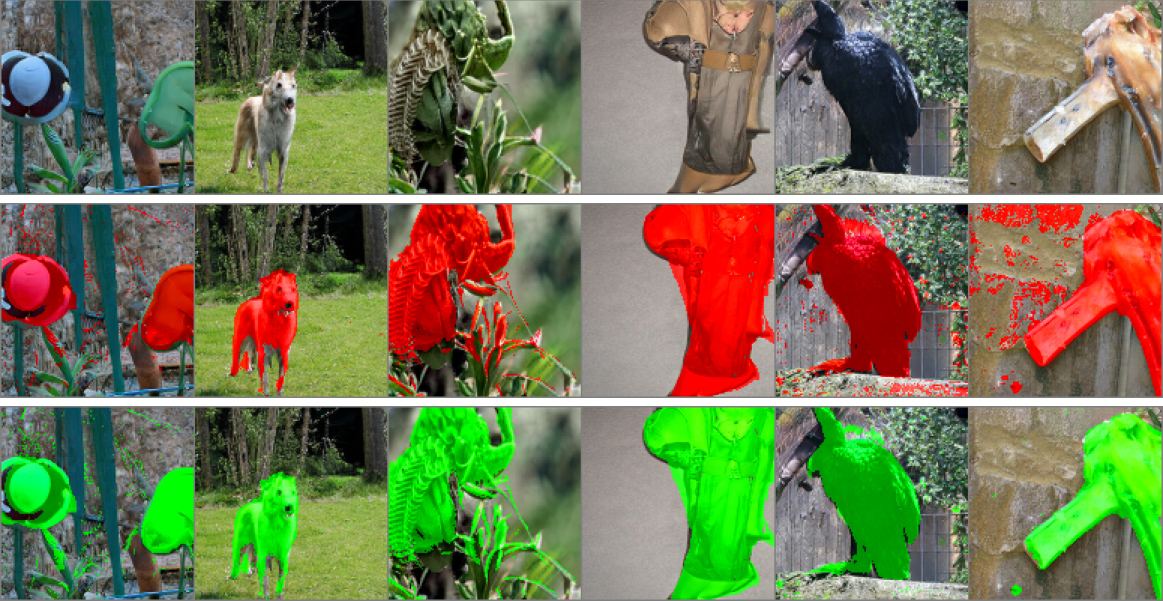}
    \caption{\textit{First row}: generated images $G(z)$; \textit{second row}: masks generated with operators $A_1, A_2$ that optimize the common objective from Equation \ref{restoration_loss}; \textit{third row}: masks generated with the operators $A_1, A_2$ independently optimized for each sample.}
    \label{fig:individual_masks}
\end{figure}

In principle, the optimization problem (\ref{L_loss}) that induces the segmentation masks could be solved independently for each generated image. We have tried to optimize the operators $A_1, A_2$ independently for each generated sample $G(z)$ during the synthetic data generation. In this case, the resulting masks appear to be almost the same as masks produced by the operators optimized for all synthetic images simultaneously. In \fig{individual_masks} we present the masks generated in the original protocol (red) and the masks generated with the per-sample optimization (green). Mutual IoU and accuracy of these masks computed over $128$ randomly generated samples is $0.86$ and $0.96$ respectfully, therefore, they are very similar. Even though the latent shift may indeed act with a different strength, the optimal operators' indices from Equation \ref{op_to_label} seem to be rather persistent. We should also note that this per-sample training protocol would require an enormous extra time as single batch generation takes approximately 30 seconds.

%% file: main_table.tex
\begin{table*}[ht!]
 \centering
 \renewcommand\arraystretch{1.3}
\setlength{\tabcolsep}{2pt}
\begin{tabular}{|c|c|c|c|c|c|c|c|c|c|}
\hline
\multirow{2}{*}{Method} &\multicolumn{3}{|c|}{ECSSD}&\multicolumn{3}{|c|}{DUTS}&\multicolumn{3}{|c|}{DUT-OMRON}\\
\cline{2-10}
& max$F_{\beta}$ & IoU & Accuracy & max$F_{\beta}$ & IoU & Accuracy & max$F_{\beta}$ & IoU & Accuracy\\
\hline
HS & 0.673 & 0.508 & 0.847 & 0.504 & 0.369 & 0.826 & 0.561 & 0.433 & 0.843\\
wCtr & 0.684 & 0.517 & 0.862 & 0.522 & 0.392 & 0.835 & 0.541 & 0.416 & 0.838\\
WSC & 0.683 & 0.498 & 0.852 & 0.528 & 0.384 & 0.862 & 0.523 & 0.387 & \textbf{0.865}\\
DeepUSPS & 0.584 & 0.440 & 0.795 & 0.425 & 0.305 & 0.773 & 0.414 & 0.305 & 0.779\\
BigBiGAN & 0.782 & 0.672 & 0.899 & 0.608 & 0.498 & 0.878 & 0.549 & 0.453 & 0.856\\
E-BigBiGAN & \textbf{0.797} & \textbf{0.684} & \textbf{0.906} & \textbf{0.624} & \textbf{0.511} & \textbf{0.882} & \textbf{0.563} & \textbf{0.464} & 0.860\\

\hline
\end{tabular}

\caption{The comparison of unsupervised saliency detection methods. For BigBiGAN and E-BigBiGAN we report the mean values over 10 independent runs.}

\label{tab:saliency}
\end{table*}

%% file: conclusion.tex
\section{Conclusion}
\label{sect:conclusion}

In our paper, we continue the line of works on unsupervised object segmentation with the aid of generative models. While the existing unsupervised techniques require adversarial training, we introduce an alternative research direction based on the high-quality synthetic data from the off-the-shelf GAN. Namely, we utilize the images produced by the BigBiGAN model, which is trained on the Imagenet dataset. Exploring BigBiGAN, we have discovered that its latent space semantics automatically create the saliency masks for synthetic images via latent space manipulations. We propose to use the BigBiGAN's encoder to fit this pipeline for a particular dataset. As shown in experiments, this synthetic data is an excellent source of supervision for discriminative computer vision models. The main feature of our approach is its simplicity and reproducibility since our model does not rely on a large number of components/hyperparameters. On several standard benchmarks, we demonstrate that our method achieves superior performance compared to existing unsupervised competitors.